# A Prompt Refinement-based Large Language Model for Metro Passenger Flow Forecasting under Delay Conditions

**Ping Huang**
College of Urban Transportation and Logistics
Shenzhen Technology University, Shenzhen, China, 518118
Email: huangping2023@email.szu.edu.cn

**Yuxin He***
College of Urban Transportation and Logistics
Shenzhen Technology University, Shenzhen, China, 518118
Email: heyuxin@sztu.edu.cn

**Hao Wang**
College of Electronics and Information Engineering
Shenzhen University, Shenzhen, China, 518118
Email: haowang@szu.edu.cn

**Jingjing Chen**
College of Urban Transportation and Logistics
Shenzhen Technology University, Shenzhen, China, 518118
Email: chenjingjing@sztu.edu.cn

**Qin Luo**
College of Urban Transportation and Logistics
Shenzhen Technology University, Shenzhen, China, 518118
Email: luoqin@sztu.edu.cn

*Ping Huang, Yuxin He\*, Hao Wang, Jingjing Chen and Qin Luo*

**ABSTRACT**
Accurate short-term forecasts of passenger flow in metro systems under delay conditions are crucial for emergency response and service recovery, which pose significant challenges and are currently under-researched. Due to the rare occurrence of delay events, the limited sample size under delay condictions make it difficult for conventional models to effectively capture the complex impacts of delays on passenger flow, resulting in low forecasting accuracy. Recognizing the strengths of large language models (LLMs) in few-shot learning due to their powerful pre-training, contextual understanding, ability to perform zero-shot and few-shot reasoning, to address the issues that effectively generalize and adapt with minimal data, we propose a passenger flow forecasting framework under delay conditions that synthesizes an LLM with carefully designed prompt engineering. By Refining prompt design, we enable the LLM to understand delay event information and the pattern from historical passenger flow data, thus overcoming the challenges of passenger flow forecasting under delay conditions. The propmpt engineering in the framework consists of two main stages: systematic prompt generation and prompt refinement. In the prompt generation stage, multi-source data is transformed into descriptive texts understandable by the LLM and stored. In the prompt refinement stage, we employ the multidimensional Chain of Thought (CoT) method to refine the prompts. We verify the proposed framework by conducting experiments using real-world datasets specifically targeting passenger flow forecasting under delay conditions of Shenzhen metro in China. The experimental results demonstrate that the proposed model performs particularly well in forecasting passenger flow under delay conditions.
**Keywords:** Passenger flow forecasting, delay event, prompt engineering, large language model

*Ping Huang, Yuxin He\*, Hao Wang, Jingjing Chen and Qin Luo*

**INTRODUCTION**

Metro, with its efficient, fast and environmentally friendly features, has become a key tool for easing urban traffic congestion and reducing environmental pollution. The operational efficiency of the metro system directly affects the daily travel of urban residents and the operational efficiency of the city. metro delays, such as those caused by equipment failures, extreme weather, traffic accidents, or large-scale social events, are frequent and unpredictable, and have a serious impact on safety, service efficiency, and the operation of the entire urban transportation system. These delay events not only disrupt the normal operation of the transportation system, but also cause fluctuations in passenger flow, which can have a significant impact on the passenger experience and transportation organization. For example, extreme weather conditions may lead to the temporary closure of multiple metro stations, major traffic accidents may cause prolonged line blockages, and large-scale social events may lead to a spike in passenger traffic in certain areas. All these scenarios require more flexibility and accuracy in traffic forecasting systems. In this context, how to accurately predict changes in metro passenger flow when delays occur, and timely adjust the operation plan and resource allocation has become a major challenge for transportation managers. The unpredictability and suddenness of delay events often make it difficult for the rail transit system to respond quickly and effectively, thus affecting the efficiency of emergency response and resource dispatching by transportation management.

In addition, in the case of a delay event, the travel behavior of passengers and the distribution of passenger flow will change significantly. Traditional passenger flow forecasting methods based on historical data are often difficult to adapt to these dynamic and complex passenger flow changes, resulting in less accurate forecasts. For example, an unexpected equipment failure may lead to the suspension of service on a major route, forcing passengers to switch to other transportation modes or routes, in which case the original passenger flow pattern will change significantly. Existing passenger flow forecasting models, which are usually based on passenger flow patterns under regular conditions, fail to adequately take into account the impact of such events on passenger flow dynamics.

Currently, passenger flow forecasting methods mainly include three categories: traditional statistical learning, traditional machine learning methods, and deep learning methods. In the early days, statistical methods such as ARIMA models (*1,2*) for time series analysis and vector autoregression (VAR) (*3*) models were widely used for their efficient handling of small datasets and their utility in short- and medium-term forecasting. These models rely on the statistical properties of historical data to predict future passenger flows, mainly by analyzing trends and seasonal patterns in the data. With the improvement of technology and computational power, traditional machine learning techniques such as Support Vector Machines (SVM) (*4*) and K-Nearest Neighbor (K-NN) (*5,6*) are beginning to be applied to passenger prediction, gaining attention for their powerful ability to handle high-dimensional data and capture complex nonlinear relationships in passenger data. These methods improve the accuracy and efficiency of prediction by learning decision boundaries or patterns from data. Moving deeper into the deep learning era, methods for passenger prediction have become more diverse and powerful. Deep learning models such as Convolutional Neural Networks (CNNs) (*7*) and Recurrent Neural Networks (RNNs) (*8*) are becoming mainstream due to their advantages in handling large-scale datasets and complex model structures. CNNs efficiently capture spatial dependencies through their convolutional layers, while RNNs optimize the processing of time series data. These models not only improve prediction accuracy, but are also able to handle the nonlinear and high-dimensional nature of traffic data. Despite the success of CNNs and RNNs in the field of passenger prediction, they still have limitations in dealing with non-Euclidean-structured data. At this point, Graph Convolutional Networks (GCNs) emerged with their unique advantages in dealing with graph-structured data (*9*). GCNs provide a new approach for modeling dynamic changes in passenger flow by propagating information among nodes and being able to capture complex spatial dependencies and network structural properties. However, GCN often encounters the problem of excessive smoothing when dealing with global spatial patterns, limiting its ability to capture long-range dependencies. To overcome these limitations, attention-based models, such as Transformer (*10*), have been introduced to passenger flow prediction. These models are able to model dynamic dependencies in time and space more flexibly through a dynamic attention mechanism

*Ping Huang, Yuxin He\*, Hao Wang, Jingjing Chen and Qin Luo*

that does not rely on a fixed adjacency matrix. This approach provides a new perspective that enables the models to effectively capture and utilize the complex spatial-temporal dependencies in passenger data on a global scale, which greatly improves the accuracy and efficiency of the predictions. However, despite the fact that Transformer models perform well in handling time series data, they still face some limitations in passenger prediction applications. First, since Transformer's self-attention mechanism requires computation for each pair of elements in the sequence, this leads to high computational complexity and memory consumption when dealing with large-scale data. In addition, Transformer usually requires a large amount of data for training, which is not always feasible in passenger prediction, especially in scenarios with sparse or incomplete data. Second, although it is effective in capturing long-term dependencies, its performance relies heavily on the tuning of parameters and the depth of the model during training, which can lead to training instability and difficult debugging issues.

To address these issues, researchers have begun to explore methods that incorporate large language models (LLMs). LLMs are able to effectively reduce the resource requirements for model training by utilizing pre-training and parameter-sharing strategies while improving the model's ability to adapt to sparse data (*11*). Furthermore, LLMs have shown potential in dealing with complex temporal and spatial dependencies by pre-training on large amounts of multi-task data. This enables them to provide deeper insights and more accurate prediction results in passenger prediction, especially when dealing with high-dimensional and multimodal passenger data.

The LLMs method utilizes Natural Language Processing (NLP) to generate predictions, providing a new way of thinking about passenger flows forecasting. By converting this numerical and categorical data into a natural language format, the method makes it possible to understand and forecast traffic flow using existing language models. This shift to language-based forecasting opens up new avenues for improving forecasting capabilities while reducing the need to design complex, specific forecasting models. The key challenge in language-based prediction methods is to design effective prompts to guide the model in understanding the task and context. In passenger flow prediction, converting numerical data of passenger flow and related auxiliary information into sentences that can be understood by the language model is a decisive factor in realizing high-precision forecasting. As a bridge between raw numerical data and natural language descriptions, the design of prompts has a direct impact on the ability of language models to capture traffic flow situations. Different prompts may cause the model to generate very different linguistic responses, thus affecting the accuracy of the forecast. The main contributions of this work are as follows:

a) A passenger flow forecasting framework combining systematic prompt engineering and LLM is proposed for addressing the few-shot dilemma in passenger flow forecastingunder delay conditions.
b) This work combines the Chain of Thought (CoT) (*12*) method, refining prompts through multiple dimensions, significantly enhancing the predictive performance of the LLM under delay conditions.
c) The experiments on real-world cases verify that the proposed model has high accuracy and reliability under delay conditions.

**RELATED WORK**

In this section, we review two aspects of related work, one on models related to passenger forecasting, and the other on LLMs.

**Models Related to Passenger Flow Forecasting**
Traditional statistical learning methods, such as ARIMA models and exponential smoothing, have long been used to work with linear time-series data. They rely on the statistical properties of historical data to predict upcoming traffic flows. By capturing the time-dependent and seasonal patterns inherent in traffic data, these methods provide a solid foundation for short- and medium-term traffic forecasting. These methods perform well in terms of stability and explanatory power, but are typically only applicable to relatively simple forecasting problems. With the development of data science, machine learning methods such as support vector machines (SVMs) and decision trees have been introduced to the field of traffic prediction, providing the ability to deal with nonlinear problems.While traditional machine learning

*Ping Huang, Yuxin He\*, Hao Wang, Jingjing Chen and Qin Luo*

models can effectively learn nonlinear patterns in data, they still have limitations such as the need for extensive feature engineering, difficulty in dealing with high-dimensional data, and the inability to effectively capture complex temporal dependencies. Deep learning models provide a powerful capability to deal with traffic prediction problems by virtue of their multilayer neural network structure. Due to their large number of layers and parametric quantities, these models are particularly well suited for extracting features from large and complex datasets, leading to excellent prediction performance. For example, RNNs and their variants, such as Long Short-Term Memory Networks (LSTMs) and Gated Recurrent Units (GRUs), and especially gated RNN structures efficiently manage long-term information, significantly enhancing the model's ability in capturing long-term dependencies in time series data. In recent years, RNNs and their variants are often used as an important component of hybrid deep neural network models and play a key role in capturing temporal patterns in traffic prediction, especially when dealing with dynamically changing and periodic events. In addition, CNNs, a deep learning method widely used in image and video analysis, have achieved great success in processing highly spatially correlated traffic data. CNNs are capable of automatically and unsupervised learning of spatial features from the input data through their convolution and pooling operations, and these features extracted directly from the raw data enable the models to identify previously unknown spatial dependencies. Further, graph neural networks (GNNs) are used to deal with spatial dependencies in non-Euclidean transportation networks, where the transportation network is naturally viewed as a graph, with stations and road segments as nodes and roads or paths connecting them as edges. This structure enables GNNs to learn complex relationships and interactions among nodes, providing deeper insights into the dynamics in urban transportation systems. In addition, the Transformer model, initially applied in the field of NLP, provides a new and powerful tool for traffic prediction; Transformer relies solely on an attention mechanism to process sequential data, and this attention mechanism enables the model to focus on different parts of the input data sequence, thus effectively capturing long-range dependencies. Subsequent research has seen deep learning models becoming more and more complex, and various model improvements in both temporal and spatial modules continue to drive traffic prediction techniques, but a problem can be identified in that they lack in generalizability as well as flexibility, and cannot be fully adapted to real-world traffic data in different domains.

**Large Language Models**
The application of LLMs in traffic prediction is an emerging research area, e.g., Transformer-based GPTs (*13*) contain billion-level parameters and extensive contextual understanding on massive datasets through pre-training on large-scale corpora, and they utilize their powerful linguistic modeling capabilities to process traffic data (e.g., traffic flow, delay information, etc.), which have made excellent progress so far. For example, TEST, proposed by Sun et al. (*14*) activates the ability of LLM to process time series data by transforming them into LLM-compatible representations, and in particular, enables LLM to better accept the embedding of these data by creating soft prompts. Similarly, TEMPO, proposed by Cao et al. (*15*) effectively learns time series representations by decomposing time series data into three components: trend, seasonality and residuals, and utilizing soft prompts to guide the model to adapt to different types of time series data. In addition, TrafficGPT, proposed by Zhang et al. (*16*) creates a novel system capable of processing and analyzing traffic data, providing insightful decision support, and assisting humans in traffic control decision making through natural language conversations by combining ChatGPT with a traffic base model. The study of Xue et al. (*17*) improves traffic control decision making based on the development of a new prompt-mining framework by developing a new language-based passenger mobility prediction. The framework employs language modeling to transform numerical data into natural language sentences to predict passenger mobility. The study breaks through the limitation of fixed templates in traditional methods and improves the accuracy and efficiency of prediction by dynamically generating and refining prompts. Further, Liang et al. (*18*) focuses on the utilization of LLMs for passenger flow prediction in the context of public events. They proposed a framework based on LLMs (LLM-MPE) to process and predict changes in pedestrian flow demand due to public events. This study first converted unstructured event descriptions obtained from online sources into a standardized format,

*Ping Huang, Yuxin He\*, Hao Wang, Jingjing Chen and Qin Luo*

and then decomposed historical mobility data into regular and event-related components. In addition, they devised a prompting strategy to guide and justify demand forecasting by LLMs taking into account historical mobility and event characteristics. However, they focused on routine and repetitive mobility patterns. The potential of LLMs to address more complex human mobility prediction tasks, especially in scenarios involving public events, remains to be explored. Although these studies have achieved good results in the area of traffic data processing using LLMs and in the area of traffic flow, they are more concerned with the prediction of routine passenger flows. The potential of LLMs to solve the problem of predicting passenger flows in delay scenarios is yet to be explored.

## METHODOLOGY

In this paper, we propose a new approach to metro passenger flow prediction under delay events using LLMs. Our goal is to develop a passenger flow prediction model that can combine delay data information with history metro passenger flow data to predict passenger flow. In the following sections, we provide a comprehensive overview of our approach. First, we describe the problem statement. Then, we present the overall framework of the model, focusing on generating and refining effective prompts, which is crucial for LLM models, As shown in **Figure 1**.

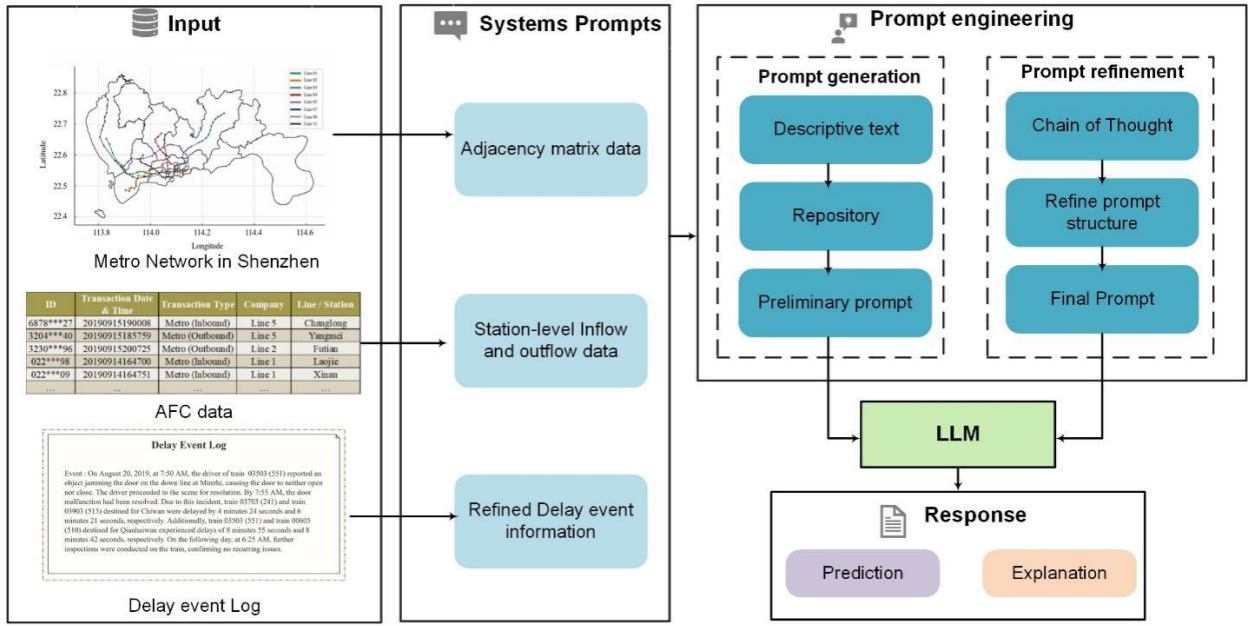

**Figure 1 Overall framework**

### Problem Statement

Metro delays refer to deviations of the metro running time from the scheduled timetable, which are caused by a variety of reasons including, but not limited to, technical failures, emergencies, weather effects, human factors, and maintenance work. The objective of this study is to use historical passenger flow information as well as information on delay events to predict access to delay-affected stations in the subway network, which is a type of spatial-temporal prediction problem. In our research framework, we focus on predicting the metro traffic flow at a future time step. Specifically, the traffic flow prediction problem can be formulated as predicting future flow values based on available historical data, which can be expressed by the following equation:

$$\{X_{t+1}, X_{t+2}, \ldots, X_{t+u}\} = F\{(X_{t-d+1}, \ldots, X_{t-1}, X_t), A, E\} \tag{1}$$

*Ping Huang, Yuxin He\*, Hao Wang, Jingjing Chen and Qin Luo*

Here, the function $F(\cdot)$ is an LLM model for passenger traffic forecasting that learns the mapping relationship between inputs and outputs and predicts the traffic attributes for the next $u$ time steps based on the flow attributes for the past $d$ time steps. $X_{t+1}, X_{t+2}, \ldots, X_{t+u}$ represent successive future predicted values and $X_{t-d+1}, \ldots, X_{t-1}, X_t$ represent continuous historical observations. $A \in \mathbb{R}^{n \times n}$ denotes the binary adjacency matrix that only contains 0 and 1, where $n$ is the number of orbital stations. $E \in \{E_T, E_L, E_H\}$ represents the subway delay event information, where $E_T$ represents the time of the delay event, $E_L$ represents the metro line where the delay event occurs, and $E_H$ represents the specific textualized representation of the subway delay event.

**Overview**
In this section, we elaborate two phases, namely the systematic prompt generation phase, and the prompt refinement phase. In the prompt generation phase, complex temporal and spatial traffic data and delay event information are transformed into a textual information format that can be understood by the model, followed by the prompt refinement phase, where we refine the prompt templates and input and output prompts through a well-designed incorporation of CoT in order to motivate step-by-step inference of the model and improve the accuracy and reliability of the predictions. The final model outputs the prediction results in JSON format.

*Systematic prompt generation phase*
In this framework, the systematic prompt generation phase is the key starting point, and the inputs to the LLM model are the first step in the prompt generation phase and are responsible for covering all the relevant information. In this study, the inputs to the model are divided into three main parts: the inputs of delay event information, the inputs of historical passenger flow data, and the inputs of the adjacency matrix. The input of delay event information mainly includes a detailed description of the comprehensive information about the train delay event, which mainly includes: the type of fault that caused the delay (train faults, signaling faults, power supply faults, and other types of faults), the specific time of occurrence (date and moment), the location (specific stations as well as line segments), and the expected scope of impact of the delay (number of stations and line segments affected). The inputs of historical passenger flow data include inflow as well as outflow at all stations on all subway lines, allowing historical passenger flow data to be collected for similar time periods as the delay event, in order to provide sufficient background information for the model as well as a basis for comparison. The inputs to the adjacency matrix reveal direct connections between stations, allowing LLM to better learn about the relationships between stations that interact with each other and their potential impact on changes in passenger flow. Since the above model inputs are not yet systematized, it is difficult for the LLM to accurately understand the task and follow up on them, so we next aggregated the above inputs into a repository of information that prepares the LLM to analyze and understand the inputs, and to generate the appropriate prompts. Subsequently, the LLM analyzes and extracts the inputs from the information database in depth, and converts the complex input data into systematic descriptive text that can be understood by the LLM. At this stage, we make it generate a large number of systematized descriptive texts through instructions to build a library of templates for generating prompts.

*Prompt refinement phase*
In this framework, the prompt refinement phase is an important step to improve the prediction performance of LLM models. In the prompt refinement phase, the main task is to fine-tune the design of the prompts by incorporating CoT to enhance the reasoning ability and prediction accuracy of LLM. The core of this phase is to refine the input prompts as well as the command output prompts in order to better guide the model for deep logical reasoning and accurate prediction.

We perform an initial screening and modification of the descriptive text created in the systematic prompt generation phase. By evaluating the logical clarity as well as the predictive power of each prompt, we can filter out the prompt words that can effectively trigger the LLM to perform more correct reasoning and prediction. In this process, we value the accuracy and completeness of each prompt in describing the

*Ping Huang, Yuxin He\*, Hao Wang, Jingjing Chen and Qin Luo*

details of the delay event, the relevance of the historical passenger flow data, and the adjacency matrix information. Next, by using CoT, we further refine each prompt to ensure that they contain sufficient intermediate logical steps that can help the model understand the complexity and multi-level factors of the problem. Inspired by Zhang et al. (*19*), we consider as much diversity as possible when constructing the CoT, thinking in multiple ways to refine the prompts. For example, for a specific delay event, the prompt first describes the basic information of the event (time, location, stations and lines affected), considers the specific time when the delay occurs as well as the historical passenger flow data for the same time period, which makes it think about how the passenger flow pattern and its correlation with the event for a specific time period. Consider the type and severity of the delay event, such as whether the delay was caused by train failure or human factors, and the duration of the delay, to make it think about the impact of the event on passenger flow. Consideration of the expected scope of impact of the delay event makes it think about whether it will cause a chain reaction affecting multiple lines or a wider area. By considering multiple dimensions in this way, the prompts will be more comprehensive and able to cover the temporal dimension as well as the spatial dimension of the passenger flow prediction problem under delays, thus guiding the model to make more accurate inference.

In addition, we will refine the prompts to reduce the bias and error of the model in prediction. This includes adjusting the level of detail of the prompts to ensure that the information is sufficient and actionable, while avoiding over-complexity, which can lead to model confusion. We also test different prompt structures in this phase, such as side-by-side prompts for each factor and nested prompts for each factor, to find the optimal prompt structure.

This phase can significantly improve the model's ability to handle complex problems through well-designed prompts and coherent logical reasoning.

*Passenger flow forecasting under delay conditions via LLM*
When predicting passenger flow, the process is mainly divided into two stages: the training process and the testing process. We train the LLM to adapt to passenger flow forecasting tasks under delay conditions. During the model training process, we continuously record the impact of different prompt templates on the accuracy of passenger flow predictions. In the testing process, we fill data information into the generated and refined prompt templates, which serve as input prompts for the LLM, in order to validate the effectiveness of the refinement methods. The results of the predictions are output in JSON format for subsequent evaluation of forecasting performance.

**EXPERIMENTAL SETTINGS AND RESULTS**
In this section, we will first describe our experiments, which include the datasets we used, the evaluation methods we employed, the baseline models we compared them to, followed by the results of our experiments.

**Dataset Description**
In order to validate our large language-based model for predicting metro passenger flow under delay events, the historical passenger flow dataset and delay event data used in this study were selected for experiments.

*Historical passenger flow dataset*
The data used in our experiments come from the AFC system of Shenzhen Metro Company in China, and the time span of Shenzhen Metro AFC swipe data is from August 5 to September 26, 2019, a period of 53 days, which includes 8 lines and 166 stations of Shenzhen Metro as of 2019. Each row of this AFC dataset represents a passenger's card swipe transaction, which includes information such as date and time of card swipe, type of entry and exit, device code, metro line, and line station, as shown in **TABLE 1**. The historical passenger flow dataset is organized, taking into account that the subway operation time is from 6:00 to 24:00 every day, we specifically process the AFC data with the corresponding time range. In this process, we perform data cleaning data filtering and data resampling, remove all outliers, and resample at

*Ping Huang, Yuxin He*, Hao Wang, Jingjing Chen and Qin Luo*

10-minute intervals to ensure the integrity of the collected passenger flow information. The historical passenger flow dataset covers 166 stations in the entire Shenzhen Metro network, and the passenger flow data, including the number of inbound and outbound passengers, is collected at 10-minute intervals.

**TABLE 1 Shenzhen Metro AFC Data Sample**

| Passenger ID | Transaction Date and Time | Type | Device Code | Line | Station | Settlement Date |
|---|---|---|---|---|---|---|
| 880023346 | 2019/8/1 4:36 | Entry | 263033104 | Line 5 | Baigelong | 20190801 |
| 881300346 | 2019/8/1 4:54 | Entry | 262016601 | Line 4 | Shenzhen North | 20190801 |

*Delayed event dataset*
The delay event dataset provides detailed information on 15 failure events that occurred from August 1, 2019 to September 30, 2019 for a variety of reasons. The delay event dataset records information about the type of fault, time of occurrence, detailed fault description, and impact on train operations for each delay event, and the **TABLE 2** shows abbreviated information about our delay event information. The text descriptions of the fault details of each delay event are more than 300 words, detailing the delays at each station during train operation, and these records are the key information we input into the model.

**TABLE 2 Delay Event Information Sample**

| Line | Delay type | Date | Time | Delay interval | Direction |
|---|---|---|---|---|---|
| Line 1 | Signaling Fault | 2019-09-19 | 18:04-19:08 | Shenzhen University-Airport East | Down |
| Line 5 | Train Fault | 2019-08-26 | 07:49-08:52 | University Town-Huangbei Ling | Up |

**Evaluation Metrics**
The most common metrics evaluated in passenger flow forecasting tasks include Root Mean Square Error (RMSE) and Mean Absolute Error (MAE). These two metrics provide quantitative measures of the performance of the prediction model, where RMSE gives more weight to large errors and MAE treats all error sizes equally. These two metrics are chosen for the performance evaluation of our experiments, with the following formulas:

$$RMSE = \sqrt{\frac{1}{n}\sum_{i=1}^{n}(y_i - \hat{y}_i)^2} \qquad (2)$$

$$MAE = \frac{1}{n}\sum_{i=1}^{n}|y_i - \hat{y}_i| \qquad (3)$$

*Ping Huang, Yuxin He\*, Hao Wang, Jingjing Chen and Qin Luo*

Where $n$ signifies the total number of samples, $y_i$ denotes the true value of passenger flow, and $\hat{y}_i$ denotes the predicted value of passenger flow.

**Baseline Models**
We have selected some representative models in the field of passenger flow prediction as Baseline Models, which are:
- **ARIMA:** A classical statistical model for time series forecasting.
- **SVR:** A machine learning method for regression analysis that uses SVM to predict continuous values.
- **LSTM:** A classical RNN variant with memory gating mechanism.
- **GC-LSTM (*20*):** A spatio-temporal prediction model combining GCN and LSTM.
- **Informer (*21*):** A variant of the Transformer model incorporating the Probabilistic Sparse (ProbSparse) self-attention mechanism.

**Case Study**
Our experiments were conducted using the well-known GPT-4 developed by OpenAI, focusing on the passenger inflow and outflow under the delay events from September 18 to September 26, 2019, due to the multiple delays that occurred during train movement in the interval during that period, and the multiple occurrences on September 26 in Shenzhen Metro Line 1 due to power supply failures as well as train failures on that day alone On September 26, there were a number of train delays on Shenzhen Metro Line 1 due to power supply failures and train breakdowns, of which Airport East Station, as the final arrival station for many of the affected trains on Metro Line 1, experienced significant delays in final arrival times due to train delays and breakdowns. In addition, our study focuses on the passenger flow forecast for Airport East Station on the morning of September 26, as the breakdowns were concentrated in the morning of September 26th. We use the historical passenger flow data and delay event information from August 5 to August 31, 2019, as the training set, and the dataset from September 1 to September 26 as the testing set.

**Results**
Our experimental prediction results are shown in **TABLE 3**, demonstrating the performance of different models in predicting passenger outflow under delayed conditions, where P1 is the prediction model for passenger flow in the prompt generation phase of our experiment, and P2 is the prediction model after our well-designed prompt refinement phase. We can directly observe that the P2 model refined by our prompts, compared with the P1 model without prompt refinement, can reflect the prediction performance improvement in both evaluation indexes, which verifies the effectiveness of the prompt refinement method. Combined with the baseline model specifically, traditional models such as ARIMA and SVR have poorer prediction performance due to the fact that these methods can only extract simple linear relationships of the data, and it is difficult to effectively capture the complex nonlinear relationships in the nonlinear passenger flow data. Based on deep learning LSTM, GC-LSTM and Informer models perform better in the prediction task compared to traditional models, especially GC-LSTM, which performs well in both RMSE and MAE, this is due to the fact that GC-LSTM can more effectively deal with spatio-temporal data with complex spatial structure and time dependence, and the traffic changes between neighboring locations often have strong correlations, and GC-LSTM is able to model such proximity effects through graph convolution operations. The prediction performance of our proposed model, the preliminary P1 model, already outperforms the conventional model in terms of RMSE and MAE, which demonstrates that the prediction method based on large language modeling has potential and is worth exploring in depth. However, the preliminary P1 model has obvious shortcomings compared to GC-LSTM, which is due to the fact that the initially generated prompt word templates are structured but not yet refined, so the prediction performance has shortcomings compared to GC-LSTM. The P2 model, which has been refined in many ways, has significantly improved its prediction performance, with the lowest RMSE, indicating that after our refinement method, it performs well in reducing the large errors and is able to make robust predictions in the face of extreme traffic variations under the influence of received delay events. Its MAE is second only to GC-LSTM, indicating that the overall prediction performance is good, but more refinement is needed.

In order to more intuitively reflect the prediction performance of each model, we do the visualization of outbound passenger flow for Airport East Station, which is seriously affected by the delay event, see **Figure 2**, and our refined P2 model has the best match with the real value curve, especially during the period of 7:30-8:30, and the P2 model accurately captures the peaks of the flow. It is able to make good predictions of traffic changes after delays at all time points.

However, our P2 also has some limitations. During the period of 6:00-6:30 on September 26, there were no passengers exiting the Airport East Station as the terminal, but our model predicted a few passengers exiting the station during this period, which is unreasonable, this phenomenon may be due to what is known as "hallucination" in LLM, which means that the output content generated by the model will be wrong or fictitious information, and this problem is also an important challenge to be faced in the subsequent research.

**TABLE 3 Performances of All Models**

| Model | RMSE | MAE |
| --- | --- | --- |
| SVR | 14.43 | 10.59 |
| ARIMA | 17.52 | 12.5 |
| LSTM | 12.56 | 9.03 |
| GC-LSTM | 12.29 | 7.76 |
| Informer | 13.46 | 8.95 |
| P1 | 12.86 | 9.14 |
| **P2** | **11.75** | **8.61** |

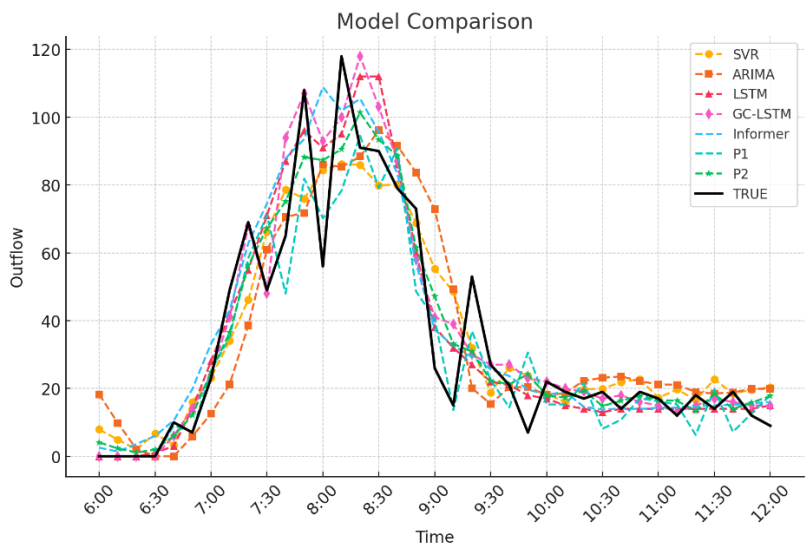

**Figure 2: Visualization of predicted passenger outbound flow at Airport East Station affected by delay events**

*Ping Huang, Yuxin He\*, Hao Wang, Jingjing Chen and Qin Luo*

## CONCLUSIONS

In this study, we propose a refinement-based passenger flow prediction method based on large language models (LLMs), especially for passenger flow under delay events. We introduce a two-stage process, including a systematic prompt generation phase and a prompt refinement phase, which enhances the model's comprehension and reasoning ability in delay situations by systematically processing complex traffic temporal and spatial data as well as delay event information. Specifically, for the prompt generation phase, this study transforms multi-source data into descriptive text that can be understood by the LLM and builds an information repository to summarize the data as well as the descriptive text. In the prompt refinement phase, we combined the Chain of Thought (CoT) method to further improve the model's prediction performance under the influence of delay events by refining the prompt words in multiple dimensions. Thereafter, by conducting comparative experiments, we verified that the performance of our passenger flow prediction method under the influence of delay events has significantly improved compared with the baseline model. Besides, the comparison experiments also reflect the model performance enhancement before and after prompt word refinement, which is a strong proof of reflecting our prompt word refinement framework. In order to further improve the ability of LLMs-based metro passenger flow prediction under delay conditions, future research can consider more information about external factors, such as large-scale events, social media data, weather data, etc., to assist the LLMs model's more comprehensive understanding ability. In addition, a more in-depth study of the design and refinement strategies of prompt words and exploring more refinement methods are challenging topics for future research.

## ACKNOWLEDGMENTS

This study was supported by the National Natural Science Foundation of China (72301180); Guangdong Basic and Applied Basic Research Foundation (2021A1515110731); Shenzhen Science and Technology Program (RCBS20231211090512002); Shanghai Key Laboratory of Rail Infrastructure Durability and System Safety (R202203); the Natural Science Foundation of Top Talent of SZTU (GDRC202126)

## AUTHOR CONTRIBUTIONS

The authors confirm their contribution to the paper as follows: study conception and design: P. Huang, Y. He; data collection: Y. He, Q. Luo; analysis and interpretation of results: P. Huang, Y. He., H. Wang; draft manuscript preparation: P. Huang, Y. He., J. Chen. All authors reviewed the results and approved the final version of the manuscript.

**REFERENCES**
1. Hamed, M.M.; Al-Masaeid, H.R.; Said, Z.M.B. Short-term prediction of traffic volume in urban arterials. Journal of Transportation Engineering 1995, 121, 249-254.

2. Ahmed, M.S.; Cook, A.R. Analysis of freeway traffic time-series data by using Box-Jenkins techniques; 1979.

3. Kamarianakis, Y.; Prastacos, P. Space–time modeling of traffic flow. Computers & Geosciences 2005, 31, 119-133.

4. Cao, J.; Fang, Z.; Qu, G.; Sun, H.; Zhang, D. An accurate traffic classification model based on support vector machines. International Journal of Network Management 2017, 27, e1962.

5. Zheng, L.; Zhu, C.; Zhu, N.; He, T.; Dong, N.; Huang, H. Feature selection-based approach for urban short-term travel speed prediction. IET Intelligent Transport Systems 2018, 12, 474-484.

6. Zhao, J.; Sun, S. High-order Gaussian process dynamical models for traffic flow prediction. IEEE Transactions on Intelligent Transportation Systems 2016, 17, 2014-2019.

7. Ma, X.; Dai, Z.; He, Z.; Ma, J.; Wang, Y.; Wang, Y. Learning traffic as images: A deep convolutional neural network for large-scale transportation network speed prediction. Sensors 2017, 17, 818.

8. Lin, G.; Ding, J.; Ding, S.; Huang, B.; Yin, Y.; Zhang, K.; Li, Z. Passenger Flow Prediction with Transformer: The Shenzhen Metro Case. In Proceedings of the 2021 International Conference on Big Data Analysis and Computer Science (BDACS), 2021; pp. 97-100.

9. Yu, B.; Yin, H.; Zhu, Z. Spatio-temporal graph convolutional networks: A deep learning framework for traffic forecasting. arXiv preprint arXiv:1709.04875 2017.

10. Cai, L.; Janowicz, K.; Mai, G.; Yan, B.; Zhu, R. Traffic transformer: Capturing the continuity and periodicity of time series for traffic forecasting. Transactions in GIS 2020, 24, 736-755.

11. Zhao, W.X.; Zhou, K.; Li, J.; Tang, T.; Wang, X.; Hou, Y.; Min, Y.; Zhang, B.; Zhang, J.; Dong, Z. A survey of large language models. arXiv preprint arXiv:2303.18223 2023.

12. Wei, J.; Wang, X.; Schuurmans, D.; Bosma, M.; Xia, F.; Chi, E.; Le, Q.V.; Zhou, D. Chain-of-thought prompting elicits reasoning in large language models. Advances in neural information processing systems 2022, 35, 24824-24837.

13. Liu, X.; Zheng, Y.; Du, Z.; Ding, M.; Qian, Y.; Yang, Z.; Tang, J. GPT understands, too. AI Open 2023.

14. Sun, C.; Li, Y.; Li, H.; Hong, S. TEST: Text prototype aligned embedding to activate LLM's ability for time series. arXiv preprint arXiv:2308.08241 2023.

15. Cao, D.; Jia, F.; Arik, S.O.; Pfister, T.; Zheng, Y.; Ye, W.; Liu, Y. Tempo: Prompt-based generative pre-trained transformer for time series forecasting. arXiv preprint arXiv:2310.04948 2023.

16. Zhang, S.; Fu, D.; Liang, W.; Zhang, Z.; Yu, B.; Cai, P.; Yao, B. Trafficgpt: Viewing, processing and interacting with traffic foundation models. Transport Policy 2024, 150, 95-105.

17. Xue, H.; Tang, T.; Payani, A.; Salim, F.D. Prompt mining for language-based human mobility forecasting. arXiv preprint arXiv:2403.03544 2024.

18. Liang, Y.; Liu, Y.; Wang, X.; Zhao, Z. Exploring large language models for human mobility prediction under public events. Computers, Environment and Urban Systems 2024, 112, 102153.

19. Zhang, Z.; Zhang, A.; Li, M.; Smola, A. Automatic chain of thought prompting in large language models. arXiv preprint arXiv:2210.03493 2022.

20. He, Y.; Zhao, Y.; Wang, H.; Tsui, K.L. GC-LSTM: a deep spatiotemporal model for passenger flow forecasting of high-speed rail network. In Proceedings of the 2020 IEEE 23rd International Conference on Intelligent Transportation Systems (ITSC), 2020; pp. 1-6.

21. Zhou, H.; Zhang, S.; Peng, J.; Zhang, S.; Li, J.; Xiong, H.; Zhang, W. Informer: Beyond efficient transformer for long sequence time-series forecasting. In Proceedings of the Proceedings of the AAAI conference on artificial intelligence, 2021; pp. 11106-11115.